\pdfoutput=1
\PassOptionsToPackage{table}{xcolor}
\documentclass{article} 
\usepackage{times}
\usepackage[accepted]{icml2025}

\usepackage{amsmath,amsfonts,bm}
\usepackage{amssymb}
\usepackage{amsopn}
\usepackage{bm} 
\usepackage{multirow}
\usepackage{flushend}
\usepackage{tabularx}
\usepackage{xspace}

\newcommand{\method}[1]{\textsc{#1}}
\newcommand{\Ours}{\method{CLOC}\xspace}
\newcommand{\Ferret}{\method{Ferret}\xspace}
\newcommand{\OWL}{\method{VESL}\xspace}
\newcommand{\Prompter}{\texttt{Prompter}\xspace}

\def\eqref#1{equation~\ref{#1}}

\def\1{\bm{1}}

\def\vl{{\bm{l}}}

\def\vp{{\bm{p}}}

\def\vx{{\bm{x}}}
\def\vy{{\bm{y}}}
\def\vz{{\bm{z}}}

\DeclareMathAlphabet{\mathsfit}{\encodingdefault}{\sfdefault}{m}{sl}
\SetMathAlphabet{\mathsfit}{bold}{\encodingdefault}{\sfdefault}{bx}{n}

\def\sL{{\mathcal{L}}}

\newcommand{\R}{\mathbb{R}}

\usepackage{hyperref}
\usepackage{url}
\usepackage{enumerate,enumitem}
\usepackage{booktabs}
\usepackage{multirow} 
\usepackage{makecell}
\usepackage[capitalize,noabbrev]{cleveref}

\renewcommand{\paragraph}[1]{\vspace{-0.5ex}\textbf{#1}}
\newcommand{\ie}{i.e.\xspace}
\newcommand{\eg}{e.g.\xspace}

\usepackage{pifont}

\definecolor{LightCyan}{rgb}{0.88,1,1}

\usepackage{tikz}
\newcommand*\circled[1]{\tikz[baseline=(char.base)]{\node[shape=circle,draw,inner sep=0pt,minimum size=9.pt] (char) {#1};}}

\usepackage{listings}
\usepackage{xcolor}

\definecolor{codegreen}{rgb}{0,0.6,0}
\definecolor{codegray}{rgb}{0.5,0.5,0.5}
\definecolor{codepurple}{rgb}{0.58,0,0.82}
\definecolor{backcolour}{rgb}{0.95,0.95,0.92}

\lstdefinestyle{mystyle}{
    backgroundcolor=\color{backcolour},   
    commentstyle=\color{codegreen},
    keywordstyle=\color{magenta},
    numberstyle=\tiny\color{codegray},
    stringstyle=\color[HTML]{a31515},
    basicstyle=\ttfamily\scriptsize,
    breakatwhitespace=false,         
    breaklines=true,                 
    captionpos=b,                    
    keepspaces=true,                 
    numbers=left,                    
    numbersep=5pt,                  
    showspaces=false,                
    showstringspaces=false,
    showtabs=false,                  
    tabsize=2,
    frame=ltb,
    framerule=0pt,
    xleftmargin=0.5em,
    xrightmargin=0em,
    framexleftmargin=0.0em,
    framextopmargin=0.5em,
    framexbottommargin=0.5em,
}

\icmltitlerunning{Contrastive Localized Language-Image Pre-Training}

\begin{document}

\twocolumn[
\icmltitle{Contrastive Localized Language-Image Pre-Training}



\icmlsetsymbol{equal}{*}

\begin{icmlauthorlist}
\icmlauthor{Hong-You Chen$^\dagger$}{comp}
\icmlauthor{Zhengfeng Lai}{comp}
\icmlauthor{Haotian Zhang}{comp}
\icmlauthor{Xinze Wang}{comp}
\icmlauthor{Marcin Eichner}{comp}
\icmlauthor{Keen You}{comp}\\
\icmlauthor{Meng Cao}{comp}
\icmlauthor{Bowen Zhang}{comp}
\icmlauthor{Yinfei Yang}{comp}
\icmlauthor{Zhe Gan}{comp}

\end{icmlauthorlist}

\icmlaffiliation{comp}{Apple AI/ML}

\icmlcorrespondingauthor{Zhe Gan}{zhe.gan@apple.com}

\icmlkeywords{}

\vskip 0.3in
]



\printAffiliationsAndNotice{$^\dagger$Work done while at Apple.} 

\newcommand{\fix}{\marginpar{FIX}}
\newcommand{\new}{\marginpar{NEW}}


\begin{abstract}
CLIP has been a celebrated method for training vision encoders to generate image/text representations facilitating various applications. Recently, it has been widely adopted as the vision backbone of multimodal large language models (MLLMs). The success of CLIP relies on aligning web-crawled noisy text annotations at \emph{image levels}. However, such criteria may be insufficient for downstream tasks in need of fine-grained vision representations, especially when understanding \emph{region-level} is demanding for MLLMs. We improve the localization capability of CLIP with several advances. Our proposed pre-training method, \textbf{C}ontrastive \textbf{Loc}alized Language-Image Pre-training (\textbf{\Ours}),  complements CLIP with region-text contrastive loss and modules. We formulate a new concept, \emph{promptable embeddings}, of which the encoder produces image embeddings easy to transform into region representations given spatial hints. To support large-scale pre-training, we design a \emph{visually-enriched and spatially-localized captioning} framework to effectively generate region-text labels. By scaling up to billions of annotated images, \Ours enables high-quality regional embeddings for recognition and retrieval tasks, and can be a drop-in replacement of CLIP to enhance MLLMs, especially on referring and grounding tasks.  
\end{abstract}

\section{Introduction}
\label{s_intro}

Vision-language (VL) pre-training has been an important foundation for the recent tremendous growth of multimodal applications. Contrastive Language-Image Pre-training (CLIP)~\citep{radford2021learning,jia2021scaling} is a successful VL representation learning method that connects images and text by contrastive training on large-scale data collected from the Web. Strong transferability and generalizability have been proven in extensive downstream tasks such as zero-shot image classification and image-text retrieval. 
Even beyond, CLIP has become arguably the default choice of vision backbone for multimodal large language models (MLLMs)~\citep{liu2023llava,mckinzie2024mm1} due to its superior prior knowledge in aligning vision and language~\citep{tong2024cambrian}.

\begin{figure*}[t!]
    \centering
    \includegraphics[width=0.85\linewidth]{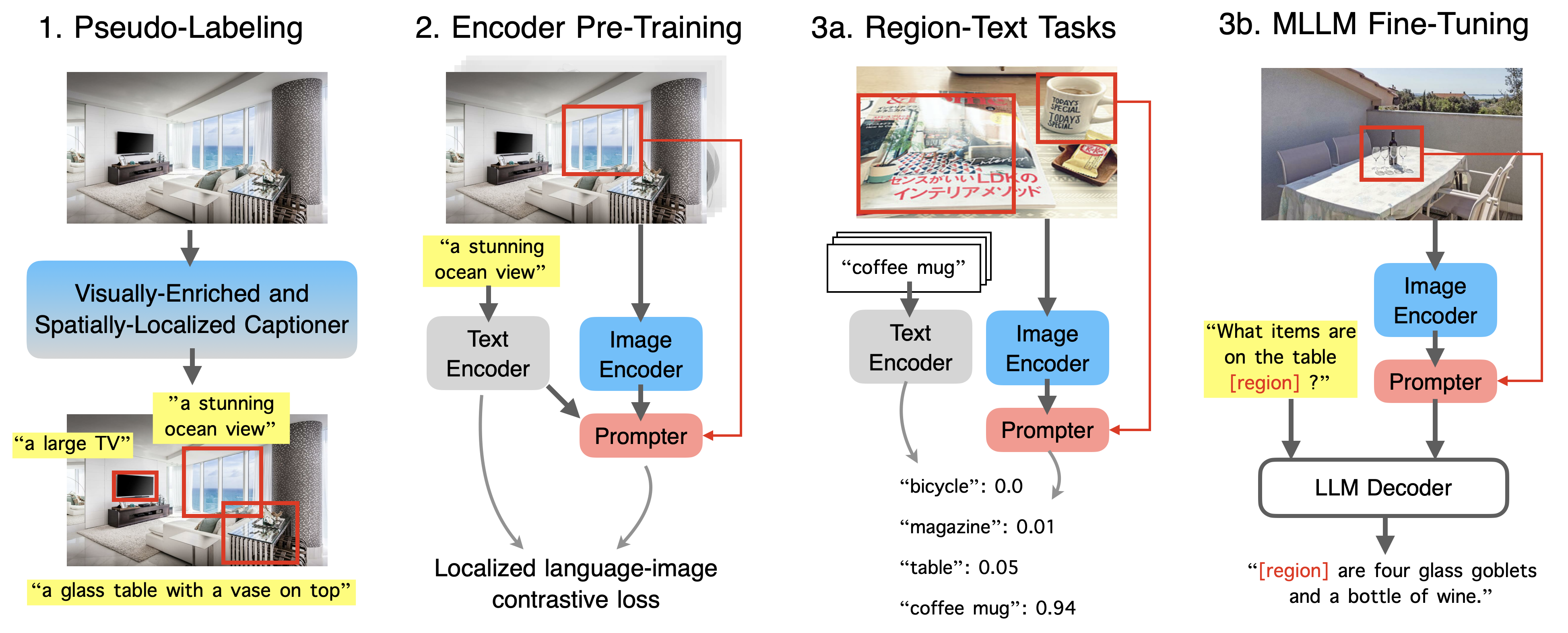}
    \vskip-10pt
    \caption{\small \textbf{Our \Ours pre-training framework.} \textbf{(1)} A visually-enriched and spatially-localized captioning pipeline pseudo-labels bounding boxes with detailed descriptions for key regions. \textbf{(2)} A lightweight \Prompter attached on the CLIP image encoder can be prompted to transform the image embedding into region-focused features. All parameters are trained end-to-end from scratch with our contrastive localized language-image loss on the annotated region-text datasets. After pre-training, \textbf{(3a)} region features can be generated via the \Prompter for region-text tasks like object classification in a training-free fashion. \textbf{(3b)} The image encoder, along with the optional \Prompter, can also strengthen MLLMs fine-tuning by enhancing their fine-grained image understanding capabilities.} 
    \label{fig:overview}
    \vspace{-3mm}
\end{figure*}

As VL research gets increasing attention, various advanced multimodal tasks are demanding stronger vision capabilities. For instance, recent MLLMs~\citep{Rasheed_2024_CVPR,ren2023pixellm,lai2023lisa,chen2023shikra,Kosmos2} have been focusing on more fine-grained understanding tasks that require comprehension of the semantics at \emph{region levels} such as visual question answering (VQA) with referring and grounding instructions. These MLLMs are fine-tuned on referring and grounding data with CLIP as the vision backbone, as seen in works like Kosmos-2~\citep{Kosmos2} and Ferret~\citep{you2023ferret,zhang2024ferret}. Due to the need for such region-level understanding, CLIP, which aligns entire images with text captions, seems insufficient, as its regular image-text contrastive loss primarily emphasizes global semantics.

To remedy such core localization capability for CLIP, we ask a challenging and fundamental question: \emph{can we pre-train a stronger image encoder (1) with enhanced localization capability that can be inherently integrated into MLLMs, (2) and refine CLIP's original image embeddings to the region level for zero-shot recognition/retrieval?}

Here, we explore a data-driven approach that complements the original CLIP image-text pre-training objective with explicit region-text supervision. Though conceptually simple, several challenges exist. \emph{First}, it lacks public datasets with region-text annotations at scales large enough for CLIP training, which typically requires hundreds of millions even billions of images. Existing region-text corpus like Visual Genome~\citep{Krishna2016VisualGC} contains about $108$K images, and the largest noisy-labeled grounded dataset GRIT~\citep{Kosmos2} features only around $20$M images. Indeed, such deficiency of labeled datasets has probably limited the literature to 
mainly consider semi-supervised or weakly-supervised approaches as somewhat a compromise~\citep{naeem2023silc,yao2022detclip,Yao_2023_CVPR}. 

\emph{Second}, a plausible solution is to scale up training data in pursuit of image regions pseudo-labeled with text annotations via some open-vocabulary detectors~\citep{minderer2024scaling,zhang2022glipv2}. Though it seems feasible, we found it non-trivial to design such a pipeline as the annotations are noisy and will greatly affect the final model performance. 
\emph{Third}, even if the region-text datasets are given, it is under-explored how to effectively train on them in terms of co-designs of training objectives, model architecture, and more design details.

To this end, we propose a new pre-training framework illustrated in~\autoref{fig:overview}, named \textbf{C}ontrastive \textbf{Loc}alized Language-Image Pre-Training (\textbf{\Ours}), to improve CLIP with better localization capability, especially for MLLMs, by overcoming the above difficulties. Our main contributions are:
\begin{itemize}[nosep,topsep=0pt,parsep=0pt,partopsep=0pt, leftmargin=*]
    \item  We propose a new learning goal,~\textbf{Promptable Embeddings}, that \emph{a strong vision encoder should produce image embeddings that can be easily transformed into region representations, given some spatial hints (\eg, box referring or text prompts)}. This formulation not only facilitates the encoder as a prior of fine-grained VL alignment, but also enables new possibilities for the interactions between the image encoder and the language decoder. 
    \item To optimize towards the goal, we design simple and minimal modifications on top of CLIP. We augment the original CLIP loss with a region-text contrastive loss, where the region embeddings are extracted from the image embedding by a lightweight extractor module conditioned on the spatial hints (\emph{\ie}, \emph{prompts}).  
    \item We design a large-scale pseudo-labeling engine to support \Ours training. We combine visual-enriched image captioners and open-vocabulary detectors for an effective recipe that improves previous practice of region annotations~\citep{minderer2024scaling,Kosmos2}. This approach yields a two-billion image-text dataset with fine-grained region-text annotations, which serves as the foundation for training our \Ours model.
    \item Through extensive experiments across 31 evaluation tasks, including standard image-text tasks, newly constructed region-text tasks, and downstream evaluations with MLLMs, we demonstrate that \Ours significantly and consistently outperforms the CLIP counterpart.
    \item We are working on releasing our pre-trained checkpoints and the constructed region-text annotations along with the final version to accelerate future research.
\end{itemize}

\section{Related Work}
\label{s_related}

\paragraph{Vision encoder pre-training.}
A popular approach to MLLMs like LLaVA~\citep{liu2023llava}, typically connects a vision encoder (\emph{\eg}, ViT~\citep{dosovitskiy2021an}) to digest visual inputs and maps them to the LLM decoder input space as token embeddings. Among various types of vision encoders~\citep{oquab2023dinov2,he2022masked}, CLIP~\citep{radford2021learning,jia2021scaling} becomes the most popular choice, due to its superior performance on MLLM benchmarks reported by recent studies~\citep{tong2024cambrian}. 

Other training approaches like captioning loss~\cite{tschannen2024image} are also popular, \citet{wan2024locca} further incorporates bounding box coordinates. However, they need training the encoder with a decoder with smaller batch sizes thus less scalable and efficient than CLIP. Also, the vision embeddings are not directly aligned with languages thus more limited for search or retrieval tasks. \emph{Therefore, our scope focuses on improving the CLIP approach}.

\paragraph{Improving localization of CLIP.}
Since CLIP was introduced, many follow-up works have been proposed to improve it from various aspects, for different target tasks, and with different approaches. 
From the aspect relevant to our work, improving the localization capability, most works specifically focus on the downstream dense vision tasks such as open-vocabulary detection~\citep{minderer2024scaling,yao2022detclip,wu2023clipself}. Another less and arguably more challenging thread is to maintain the generalizability of CLIP on image-level tasks while improving localization. Recent works~\citep{naeem2023silc,bica2024improving,dong2023maskclip} combine localization-enhancing unsupervised objectives with the CLIP loss, but do not attempt with supervision on large-scale explicit pseudo-labeled data like ours and are more computation overhead. Alpha-CLIP~\citep{Sun_2024_CVPR} shows that the SAM~\citep{kirillov2023segment} can provide useful conditions for CLIP. 

\paragraph{Synthetic annotations for pre-training.}
The literature has been exploring scalable ways to generate high-quality synthetic annotations for CLIP. For instance, several works demonstrate that visually-enriched image captions improve CLIP~\citep{lai2023scarcity}. MOFI~\citep{wu2023mofi} augments CLIP with an extreme multi-classification task. However, these works only consider image-level annotations but not explicit region-level labels. In the context of dense vision tasks like open-vocabulary detection and segmentation, pseudo-labeling in a self-training paradigm has proven an effective approach~\citep{kirillov2023segment,minderer2024scaling}. We are inspired by these efforts and combine them to enhance CLIP's localization capabilities. Our approach is promising, since the advance of these labeling methods can further improve ours in the future.

\section{\Ours}
\label{s_method}

\subsection{From Image-Text to Region-Text Alignment}
\label{ss_pre}
CLIP~\citep{radford2021learning} contrastively aligns the embedding from a pair of image and text encoders ($f_I$ and $f_T$). Let a mini-batch of $N$ image-text pairs $\{(\vx_i, \vy_i)\}_{i=1}^N$ be sampled from the large-scale training set during each training iteration. The contrastive loss is defined as follows:
{\small
\begin{align}
    \sL_{\text{CLIP}} & :=( \sL_{I\rightarrow T} + \sL_{T\rightarrow I}) / 2.\: \label{eq:clip} \\ \nonumber
        \sL_{I\rightarrow T} 
     & := -\frac{1}{N}\sum_{i=1}^N \log  \frac{\exp\big(\texttt{sim}(f_I(\vx_i), f_T(\vy_i))/\tau\big)}{\sum_{j=1}^{N} \exp\big(\texttt{sim}(f_I(\vx_i), f_T(\vy_j))/\tau\big)},
\end{align}
}
where $\texttt{sim}(\cdot,\cdot)$ is the similarity function and $\tau$ is a (learnable) temperature. The CLIP loss $\sL_{\text{CLIP}}$ averages the symmetrical contrastive loss in which cross-entropy normalized along image-to-text and text-to-image axes, respectively. 

Conceptually, $\sL_{\text{CLIP}}$ aligns images with their associated text, but it overlooks subimage semantics. We propose augmenting this with \emph{region-text} alignment on top of $\sL_{\text{CLIP}}$. Specifically, assume an image-text pair $(\vx, \vy)$ can be decomposed into image regions $\vx^{(1)},\dots, \vx^{(m)}$, and there exist regional captions $\vy'^{(m)}$ that describe the corresponding regions $\vx^{(m)}$. Thus, the original input $(\vx, \vy)$ becomes region-text pairs $\{(\vx^{(1)}, \vy'^{(1)}),\dots,(\vx^{(m)}, \vy'^{(m)})\}$, and $(\vx, \vy)$ is a special case when the ``region'' itself is the whole image. We identify several research questions and will answer them in the following sections:
\begin{enumerate}[nosep,topsep=0pt,parsep=0pt,partopsep=0pt, leftmargin=*]
    \item Considering the goal is to train an image encoder $f_I$ with enhanced localization capability, how should we formulate a region-text alignment goal that improves $f_I$? We propose a novel learning task called \emph{promptable embeddings} in Section~\ref{ss_promptable}.
    \item How to properly extract region embedding from $f_I(\vx)$ as an effective joint design? We propose a lightweight promptable region extractor in Section~\ref{ss_cloc_arch}.
    \item How to generate meaningful image regions with high-quality captions? Furthermore, in many cases, the ideal region caption $\vy'^{(m)}$ may not exist in the image-level caption, \emph{\ie}, $\vy'^{(m)}$ might not be a substring of the original $\vy$. We design an effective and scalable data engine as a visually-enriched and spatially-localized labeler to generate high-quality region-text pairs in Section~\ref{s_data}. 
    \item With the above considerations, we discuss how to train the model with minimal conflicts towards a drop-in replacement of the CLIP model in Section~\ref{ss_cloc_design}.
\end{enumerate}

\begin{figure*}[t!]
    \centering
    \includegraphics[width=0.95\linewidth]{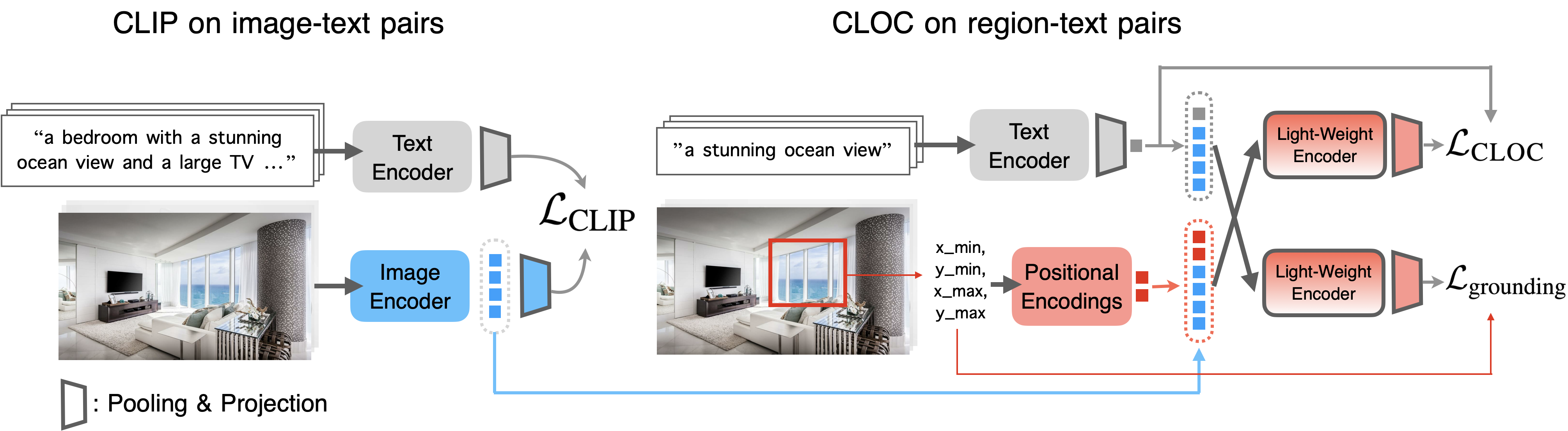}
    \caption{\small \textbf{\Ours promptable embedding architecture.} \Ours builds upon the image embedding from CLIP (before pooling and projection) and transforms it into a region-aware vision embedding given an encoded prompt; \emph{\eg}, positional encodings of box coordinates or regional caption encoded by the CLIP text encoder.}
    \label{fig:arch}
    \vskip-5pt
\end{figure*}

\subsection{Promptable Embeddings}
\label{ss_promptable}
To optimize CLIP with better feature localization and eventually learn an enhanced CLIP vision encoder $f_I$ for various VL downstream tasks, we argue that it will require at least two capabilities. ($i$) First, the encoder should recognize fine-grained small objects (\emph{\eg}, this image crop is an ``airplane wheel''). ($ii$) Second, the \emph{image embedding} produced by the encoder provides a holistic understanding such that an MLLM can reason more advanced spatial hierarchy relationships within the scene (\emph{\eg}, ``The plane is lowering its front landing gear.''). As discussed in Section~\ref{s_related}, many previous works improve CLIP toward object detection tasks thus mainly focusing on ($i$) only; \emph{\eg}, RegionCLIP~\citep{zhong2022regionclip} that crops out image regions and uses them as additional input images to re-train the CLIP encoders for recognizing objects. However, to support comprehensive VL tasks, ($i$) is necessary but insufficient without ($ii$).  

To achieve this, we introduce a new concept, \emph{promptable embedding}. We consider a scenario similar to MLLM use cases, where answers are generated using CLIP image tokens alongside a question.  We hypothesize that a strong encoder for MLLMs should produce an \emph{image} embedding that can \emph{easily be transformed into region representations, given location cues}.

We re-formulate the CLIP loss based on image-text pairs $(\vx, \vy)$ into a \textbf{localized} language-image contrastive loss for region-text alignment based on triplets of $(\{\vl\}, \vx, \vy)$, where $\vl$ is a location representation such as a bounding box, and possibly there are several boxes as a set $\{\vl\}$ per image. To make it compatible with CLIP training, we construct a promptable embedding transform module, or in short, \emph{region prompter} $\vz = \texttt{Prompter}(\vl, f_I(\vx))$, that extracts the region embedding specified by $\vl$ from the image embedding $f_I(\vx)$. This formulation is inspired by the success of the segmentation model SAM~\citep{kirillov2023segment} which predicts \emph{segmentation masks} conditioned on location prompt (\emph{\eg}, a box), while \Ours predicts a \emph{region embedding} conditioned on $\vl$ instead.  

To this end, we decompose the location-image-text triplets as localized region-text pairs. Let $\vz_{i}^{(m)} = \texttt{Prompter}(\vl_{i}^{(m)}, f_I(\vx_i))$ and $\vy_{i}^{(m)}$ is the caption of the region specified by $\vl_{i}^{(m)}$. $\vl_i^{(m)} \in \R^4$ is the $m$-th box of image $i$ represented as two coordinates (\emph{\ie}, top-left and bottom-right corners). We then formulate a symmetric region-text contrastive loss similar to~\autoref{eq:clip}:
{\small
\begin{align}
\label{eq:cloc}
    &\sL_{R\rightarrow T} := -\frac{1}{MN}\sum_{i=1}^N \sum_{\vl_{i}^{(m)} \sim \{\vl_i\}} \log \vp\\
    &\vp=\frac{\exp\Bigl(\texttt{sim}\bigl(\vz_{i}^{(m)}, f_T(\vy_{i}^{(m)})\bigr)/\tau\Bigr)}{\sum_{j=1}^N \sum_{\vl_{j}^{(m)} \sim \{\vl_j\}} \exp\Big(\texttt{sim}\bigl(\vz_{i}^{(m)}, f_T(\vy_{j}^{(m')})\bigr)/\tau\Big)},\nonumber
\end{align}
}
where $M$ is the number of regions $\vl_{i}^{(m)}$ sampled randomly per image. We set $M=4$ by default. We will discuss implementing the \texttt{Prompter} in Section~\ref{ss_cloc_arch}, and generating $\vl_{i}^{(m)}$ with $\vy_{i}^{(m)}$ in~Section \ref{s_data}. $\sL_{T\rightarrow R}$ is the symmetric contrastive loss normalized along text-to-region axis, just like in~\autoref{eq:clip}. We define $\sL_{\text{\Ours}} =( \sL_{R\rightarrow T} + \sL_{T\rightarrow R}) / 2$. 

As the \Prompter is a simple transformer encoder, it allows flexible types of prompts besides bounding boxes we have used, such as points, free-form referring, text, and etc. We further consider the case where the prompt is free-form text, and leave others for future study. We add a grounding loss that extracts a region feature from the image (\emph{\eg}, a picture of the bedroom) given its regional caption (\emph{\eg}, ``a large TV''), and predicts the bounding box with an MLP regression head, \emph{\ie},
{\small
\begin{equation}
    \sL_\text{grounding}:=\frac{1}{4MN}\sum_{i=1}^N \sum_{\vl_{i}^{(m)} \sim \{\vl_i\}} \|\vl_i^{(m)} - \texttt{Head}\big(\vz(\vy_i^{(m)})\big)\|_2,
    \label{eq:grounding_loss}
\end{equation}
}
where $\vz(\vy):= \texttt{Prompter}(f_T(\vy), f_I(\vx))$ is the grounded embedding conditioned on the text (encoded by the CLIP text encoder). All the learnable parameters are trained end-to-end. With a scalar $\lambda$, the overall loss is
\begin{equation}
    \sL := \sL_{\text{CLIP}} + \lambda(\sL_{\text{\Ours}} + \sL_\text{grounding}).
    \label{eq:loss}
\end{equation}

\subsection{\Ours Model Architecture}
\label{ss_cloc_arch}
We implement the promptable embedding introduced in~Section \ref{ss_promptable} with minimal extra modules on top of the original CLIP. As illustrated in~\autoref{fig:arch}, the CLIP model remains the same for computing $\sL_{\text{CLIP}}$. For $\sL_{\text{\Ours}}$/$\sL_\text{grounding}$, the image embedding is re-used from the CLIP ViT but before the pooled projection and normalization $f_I'$. To extract the region embedding $\vz = \texttt{Prompter}(\vl, f_I'(\vx))$ from the image, we consider the location representation $\vl$ as two coordinates (top-left and bottom-right corners of a box), each vectorized by positional encoding. The \texttt{Prompter} is a lightweight one-layer transformer encoder. It takes the positional encodings prepended with the image tokens from ViT together as the input, and outputs the region embedding with a pooled projection layer. For the grounding loss, we re-use the same CLIP text encoder for encoding the region captions $\vz = \texttt{Prompter}(f_T(\vy), f_I'(\vx))$ to predict the bounding boxes with a two-layer MLP head.  Overall, \Ours only adds lightweight additional parameters of the heads. Note that, the main overheads during forward are from ViT image encoding -- \Ours reuses it for multiple prompts.

\begin{figure*}[t!]
    \centering
    \includegraphics[width=0.8\linewidth]{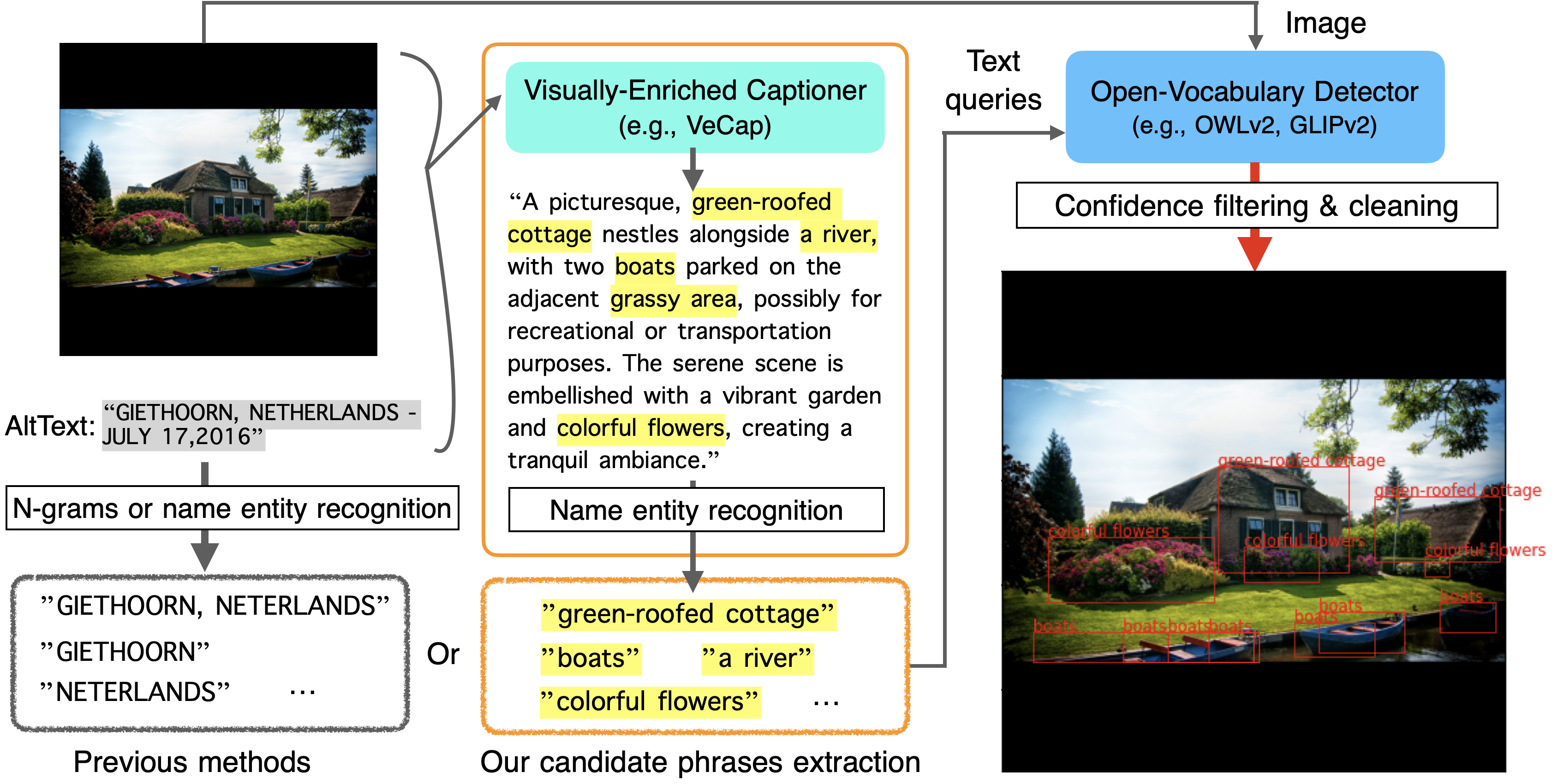}
    \vskip-5pt
    \caption{\small \textbf{Our Visually-Enriched and Spatially-Localized (\OWL) captioning pipeline.} We leverage an existing open-vocabulary detector (\emph{\eg}, OWLv2) that predicts bounding boxes on the images and assigns the labels from the given text phrase candidates. Previous methods often use the alt-text attached to the images, which is prone to insufficient region descriptions. We found it crucial to re-caption images with the visually-enriched captioner VeCap~\citep{lai2023scarcity} for better visual concept exploitation for the detector.}
    \label{fig:vesl}
    \vskip-10pt
\end{figure*}

\subsection{Discussions on Design Choices and Extensions}
\label{ss_cloc_design}
We provide discussions here on the rationale behind our design choices and some minor extensions. 

\paragraph{Extracting region embedding with visual prompts.}
Training with $\sL_{\Ours}$ in~\autoref{eq:loss} requires extracting region embeddings from the image features given the bounding boxes. Another alternative could be Region-of-Interest (RoI) pooling/alignment~\citep{he2017mask} from the spatial image feature of ViT before pooling. RoI operations are popular, especially in the object detection literature. However, as will be evidenced by worse performance in~Section \ref{s:exp}, we found spatial pooling an over-strong premise for \Ours pre-training here for several reasons. 

First, unlike object detection datasets that typically contain golden labels, here the pseudo-labels are much noisier on the large-scale web-crawled images. The resulting RoI features may be inaccurate due to the imprecise bounding boxes, making model training less effective.
Second, unlike dense vision tasks that directly rely on the spatial features, MLLM has a transformer decoder that consists of several attention layers such that the constraint of semantics in the spatial feature space becomes somewhat indirect. Our \Prompter mimics such inductive bias in pre-training via a single-attention-layer encoder that may leverage better global context reasoning compared to RoIs.

\paragraph{Avoiding region-text conflicts.}
While region annotations introduce location information, a concern of contrastive learning may be similar objects within an image (\emph{\eg}, ``boats'' in the harbor) or a mini-batch. To mitigate it, we apply two tricks. First, fortunately, we found it sufficient to sample a few regions per image for each update, \emph{\eg}, we set $M=4$ in~\autoref{eq:cloc} in experiments. Second, we can filter similar texts when computing the negatives in the contrastive loss. More specifically, we ignore the pairs of $\big(\vz_{i}^{(m)}, f_T(\vy_{j}^{(m')})\big)$ in the denominators of both~$\sL_{R\rightarrow T / {T\rightarrow R}}$, if $\texttt{sim}\big(f_T(\vy_{i}^{(m)}), f_T(\vy_{j}^{(m')})\big) > 0.9$, without gradients on $f_T$.

\section{\OWL Captioning Pipeline}
\label{s_data}

\begin{table}[t]
    \minipage{1\linewidth}
    \vskip -5pt
    \caption{\small \textbf{Region-text dataset statistics.} We summarize the text token length for both images and regions. Partial statistics of the proprietary datasets revealed by their papers.  $^*$The $20$M subset of GRIT is released; we removed the invalid URLs.}
    \label{tab:data}
    \centering
    \scriptsize
    \setlength{\tabcolsep}{1pt}
     \renewcommand{\arraystretch}{1}
    \begin{tabular}{l|cccc}
    \toprule
    Dataset & \parbox{1cm}{\centering \# of images} & \parbox{1cm}{\centering regions per image} & \parbox{1cm}{\centering image text len.} & \parbox{1cm}{\centering region text len.}\\
    \midrule
    Flickr Entities~\citep{plummer2015flickr30k} & 32K & 8.7 & -- & --\\
    RefCOCO~\citep{yu2016modeling} & 20K & 2.5 & -- & 3.6 \\
    RefCOCO+~\citep{yu2016modeling} & 20K & 2.5 & -- & 3.5\\
    RefCOCOg~\citep{mao2016generation} & 27K & 2.1 & --  & 8.4\\
    Visual Genome~\citep{Krishna2016VisualGC} & 108K & 38.0 & -- & -- \\
    \midrule
    GRIT (prop.)~\citep{Kosmos2} & 91M & 1.5 & -- &  4.7\\
    GRIT (released)~\citep{Kosmos2}$^*$ & 17M & 1.8& 17.2 & 4.6\\
    Florence-2 (prop.)~\citep{xiao2024florence} & 126M & 5.4 & 70.5 & 2.6\\
    OWLv2 (prop.)~\citep{minderer2024scaling} & 2B  & -- & -- & --\\ 
    \midrule
    WiT labeled w/ \cite{minderer2024scaling} & 300M & 5.1 & 17.1 & 3.9 \\
    \OWL WiT (Ours) & 300M & 11.6 & 44.9 & 2.1\\
    \OWL WiT+DFN (Ours) & 2B & 11.5 & 35.9  & 2.1\\
    \bottomrule
    \end{tabular}
    \endminipage
    \vskip -15pt
\end{table}

As discussed in Section~\ref{s_intro} and~\ref{ss_pre}, a key bottleneck of \Ours is the region-text datasets in terms of both the data size and the label quality, since there are no public datasets with region-text annotations at scales large enough for contrastive pre-training. Inspired by recent works that enrich image captions for better CLIP training, we make a step further for \textbf{V}isually-\textbf{E}nriched and \textbf{S}patially-\textbf{L}ocalized (\textbf{\OWL}) captioning that generates fine-grained captions at \emph{region level}. The goal of \OWL is, given an image with the original web-crawled alt-text, annotates it with the grounded bounding boxes each associated with a caption in natural language for optimizing~\autoref{eq:cloc} in Section~\ref{ss_promptable}.

Concretely, \OWL is a pseudo-labeling pipeline with the following steps, with pseudo codes in~\autoref{sup-s-vesl}:

    1. \textbf{Re-captioning with visual concept exploitation}: We follow the VeCap2~\citep{lai2023scarcity,lai2024revisit} to generate long, diverse, and detailed image captions. 
    
    2. \textbf{Region phrase candidates extraction}: We apply name entity recognition (NER) to extract phrases from the visually-enriched captions as candidate describing a region inside the image, inspired by~\citet{zhang2022glipv2}. 
    
    3. \textbf{Open-vocabulary detection with extracted phrases}: the final region-text annotations are generated by a pre-trained open-vocabulary detector. It matches the phrases extracted from Step 2 to the bounding boxes proposed by the detector. We adopted OWLv2 detector~\citep{minderer2024scaling} which combines the CLIP image/text encoders with detection heads. The boxes with confidence larger than $0.1$ are kept as the region location and the most confident phrases are considered as their captions.

\paragraph{Remarks.} We highlight our insights behind the proposed recipe. The most relevant work was proposed in~\citep{minderer2024scaling} that scales up open-vocabulary (OV) detection via self-training. We are inspired by its success and extend it to \Ours contrastive learning with important modifications. Different from~\citep{minderer2024scaling} that generates candidate phrases from the $n$-grams of the web-crawled alt-text of the images for OV detection, we found the alt-text might not have enough details describing the image region content, thus limiting the diversity and quality of the annotations predicted by the OV detector. We thus caption each image augmented with more visual details. However, the long captions make the $n$-grams candidates verbose and grow exponentially, thus we generate high-quality candidates via name entity recognition instead. We found such a pipeline produces training data more suitable for \Ours, as will be validated in Section~\ref{s:exp}.

\paragraph{Our pre-training datasets.} Our pre-training data consists of two parts: ($i$) image-text pairs, and ($ii$) region-text pairs. For image-text pairs, we reproduce the image re-captioning pipeline from VeCap~\citep{lai2023scarcity}, and generate synthetic captions for WiT-300M~\citep{wu2023mofi} and DFN-5B~\citep{fang2023data} images.
For region-text pairs, we pseudo-label WiT-300M and a 2B-image subset of DFN-5B using our \OWL pipeline. In \OWL, we adopted the official OWLv2 L/14 model~\citep{minderer2024scaling} as the open-vocabulary detector. All images are pseudo-labeled with $448\times448$ resolution, where a maximum number of $20$ phrase queries are sampled for moderate computation budget.~\autoref{tab:data} summarizes the statistics of existing region-text datasets and ours. Notably, we also ablate annotating WiT-300M following~\cite{minderer2024scaling} and found it detects \emph{less} objects with longer region text, likely due to verbose $n$-grams of alt-text are in lower quality than our approach, as discussed in the remarks.

\begin{table*}[t]
	\caption{\small Zero-shot evaluation on image-level tasks (recall@1 of COCO retrieval and accuracy of ImageNet (IN) classification) and region-level tasks (recall@10 of GRIT region retrieval and mAcc of region object recognition on COCO and LVIS), using ViT-B/16 as the default encoder backbone. The indentation with different symbols denotes removing (--) or changing a component ($\circ$). 
	}
	\label{tbl:zero-shot}
        \scriptsize
	\centering
	\setlength{\tabcolsep}{0.85pt}
        \renewcommand{\arraystretch}{0.5}
		\begin{tabular}{llcc|cccc|cccc|cc}
			\toprule
			 & Models & \multicolumn{2}{c|}{Training Labels} & \multicolumn{4}{c|}{\centering Image tasks} & \multicolumn{4}{c|}{\centering Region tasks} & \multicolumn{2}{|c}{\centering Avg.}
                \\
                \midrule
                &  & Image & Region & \parbox{1.5cm}{\centering COCO-i2t}  & \parbox{1.5cm}{\centering COCO-t2i} & INv1 & INv2 & \parbox{1.2cm}{\centering GRIT-r2t} & \parbox{1.2cm}{\centering GRIT-t2r} & \parbox{1.5cm}{\centering RecCOCO}  & \parbox{1.5cm}{\centering RecLVIS} & Image & Region \\ 
                \midrule
                \midrule
			\circled{1}  & OpenAI-CLIP & {\scriptsize proprietary} & - & 52.4 &	33.1&	68.3&	62.3 & -& -& -& -&54.0&-\\
			\circled{2}  & CLIP & {\scriptsize WiT+DFN} & - & 66.3	&45.1	&76.2	&69.6
& -& -& -& -&64.3&-\\
			\midrule
                \rowcolor{LightCyan}
			\circled{3} &  \Ours  & WiT & WiT & 68.8&	50.1&	66.7&	59.7&	65.1&	67.2&	70.6&	26.7&	61.3&	57.4\\
			\circled{4}  &  \quad -- \Prompter & WiT& WiT & 67.0&	49.7&	65.6&	58.6&	44.8&	4.4&	55.3&	13.2&	60.2&	29.4\\
			\circled{5}  &  \quad  -- \OWL & WiT& WiT & 53.9&	36.3&	66.6&	59.5&	71.5&	63.8&	62.2&	22.2&	54.1&	54.9\\
			\circled{6} & \quad $\circ$ w/ GLIPv2 & WiT& WiT & 68.8&	50.0&	65.8&	59.2&	67.9&	71.1&	64.9&	23.1&	61.0&	56.8\\
			\midrule
			\rowcolor{LightCyan}
			\circled{8}  &  \Ours & {\scriptsize WiT+DFN} & WiT & 66.1&	46.5&	75.5&	68.6&	65.8&	67.4&	70.1&	27.2&	64.2&	57.6 \\
			\circled{9}  &  \quad -- \Prompter & {\scriptsize WiT+DFN} & WiT & 65.8&	46.5&	75.7&	68.0&	55.5&	18.4&	67.1&	24.6&	64.0&	41.4\\
			\circled{10} &  \quad -- text filtering & {\scriptsize WiT+DFN} & WiT & 65.4&	46.0&	75.7&	68.4&	66.3&	66.5&	68.7&	24.8&	63.9&	56.6\\
			\circled{11} &  \quad -- $\sL_{\text{grounding}}$ & {\scriptsize WiT+DFN} & WiT & 66.0&	46.3&	75.7&	67.9&	66.0&	66.8&	70.0&	25.8&	64.0&	57.2\\
                \circled{12} &  \quad $\circ$ $M=2$  & {\scriptsize WiT+DFN} & WiT & 66.6&	46.2&	75.5&	67.9&	66.5&	67.0&	69.8&	25.8&	64.1&	57.3\\
                \midrule
                \rowcolor{LightCyan}
                \circled{13} & \Ours & {\scriptsize WiT+DFN} & {\scriptsize WiT+DFN} & 69.2&	49.3&	74.9&	67.0&	63.9&	65.9&	71.1&	28.5&	65.1&	57.3\\
                \circled{14} & \quad -- \Prompter & {\scriptsize WiT+DFN} & {\scriptsize WiT+DFN} & 70.2&	49.7&	74.7&	67.6&	65.7&	23.0&	67.1&	25.4&	65.6&	45.3\\
                \circled{15} & \quad -- \OWL & {\scriptsize WiT+DFN} & {\scriptsize WiT+DFN} & 65.3&	46.6&	75.5&	67.7&	55.7&	22.3&	66.3&	25.3&	63.8&	42.4\\ 
                \midrule
                \midrule
                \circled{16} &  \quad $\circ$ ViT L/14 & {\scriptsize WiT+DFN} & {\scriptsize WiT+DFN} & 74.8&	54.4&	80.1&	73.2&	66.9&	68.3&	72.9&	32.6&	70.6&	60.2\\
                \circled{17} &  \quad $\circ$ ViT H/14 & {\scriptsize WiT+DFN} & {\scriptsize WiT+DFN} & 75.7&	55.1&	81.3&	74.7&	67.4&	69.4&	73.0&	35.6&	71.7&	61.3\\
			\bottomrule
		\end{tabular}
  \vskip-10pt
\end{table*}

\section{Experiments}
\label{s:exp}

\subsection{Setup \text{(more details in~\autoref{sup-s-exp})}}
\label{ss_setup}
\paragraph{Pre-training.}
We follow OpenAI-CLIP~\citep{radford2021learning} to train both our CLIP baseline model and \Ours model using a similar budget of around 14B images seen. For a fair comparison, we use the same hyper-parameters and images for both the CLIP baseline and \Ours. We experimented with the ViT B/16 and L/14 architectures, pre-trained with $224\times 224$ and $336\times 336$ image resolutions, respectively. All parameters are trained end-to-end from scratch. We implement the in JAX~\citep{jax2018github}. 

\paragraph{Evaluation tasks.} The image encoders are evaluated across a wide range of downstream tasks. First, we assess performance on ImageNet image classification~\citep{deng2009imagenet,pmlr-v119-shankar20c} and COCO retrieval~\citep{lin2014microsoft}. Second, we construct region-level tasks, including COCO object recognition and region-text retrieval using the GRIT dataset~\citep{Kosmos2}. Furthermore, we show \Ours is particularly useful for MLLMs, validated by the Ferret model~\citep{you2023ferret} which requires fine-grained image understanding for referring and grounding tasks. We also evaluate on general multimodal benchmarks using LLaVA-1.5~\citep{liu2023llava} and Open-LLaVA-NeXT~\citep{liu2024llavanext,chen2024open}, which both use the 7B Vicuna LLM. For all evaluation tasks, we use the same official hyper-parameters, fine-tuning datasets, and codebase for all the image encoders we experimented with, without specific tuning. 

\subsection{Image and Region Classification and Retrieval}
\label{ss_zeroshot}

\Ours encoder produces not only image embedding but also region embeddings. It can be used directly for region-level tasks without further training, in analogy to the zero-shot capability of CLIP on images. This emergent capability enables us to construct region-level zero-shot tasks for fast development and ablation study. 

In addition to \emph{image}-level evaluation like ImageNet classification and COCO image-text retrieval, we additionally construct \emph{region}-level tasks, including region object recognition and text retrieval. More specifically, the region-level tasks leverage the labeled bounding boxes in the evaluation set for \Ours to extract region embedding. For region retrieval, we use a validation set of the GRIT dataset~\citep{Kosmos2} and encode both the image regions and the region captions. For region classification, the class names are encoded as text embedding ($80$ / $1203$ classes for COCO / LVIS, respectively), and the highest cosine similarity for each region embedding is predicted as its class. We highlight important variables for the performance in~\autoref{tbl:zero-shot} with the following observations:
\begin{itemize}[nosep,topsep=0pt,parsep=0pt,partopsep=0pt, leftmargin=*]
    \item \Ours performs decently on region-level tasks\footnote{For reference, in a different setup,~\citet{wu2023clipself} reports 46.5\% mAcc on the COCO region classification task, trained with $320 \times 320$ COCO images directly. In contrast, our approach achieves over 70\% mAcc, pre-trained on a $224 \times 224$ web-crawled dataset with our object labels (thus not a fair comparisons).} with strong image-level performance (\circled{2} vs.\ \circled{8} \circled{13}).
    
    \item The \Prompter is an important ingredient for \Ours's success to go beyond CLIP (\circled{3} \circled{8} \circled{13} vs.\ \circled{4} \circled{9} \circled{14}). We replace the \Prompter with RoI alignment to extract region features and train with $\sL_\text{\Ours}$ (similar to~\cite{shi2025umg}). We found it performs much worse on region-level tasks, possibly due to difficulties of strong RoI constraints and the noisy labels as discussed in~Section \ref{ss_cloc_design}.
    
    \item \OWL helps, as the visually-enriched captions improve image retrieval tasks (as expected~\citep{lai2023scarcity}) and the versatile visual concepts candidates facilitate the OV detector, supporting~Section \ref{s_data} (\circled{3} \circled{13} vs.\ \circled{5} \circled{15}). 
    
    \item OV detector OWLv2 $>$ GLIPv2 in \OWL (\circled{3} vs.\ \circled{6}). 
    
    \item Tricks in~Section \ref{ss_cloc_design} offer slight performance gains, but $\sL_\text{\Ours}$ is already highly effective on its own (\circled{10} \circled{11}).
    
    \item Practically, sampling $2$/$4$ boxes works (\circled{12}) well already. 
    
    \item Scaling up images saturated on region tasks but further improved on MLLM tasks (\circled{3} \circled{8};~\autoref{tab:ferret-bench}.
    
    \item Scaling up the ViT model sizes can further improve both image and region tasks (\circled{13} \circled{16} \circled{17}). 
\end{itemize}

Overall, \Ours not only achieves strong performance on image-level tasks, but unlocks zero-shot region-level capability. Below, with our design choices validated, \circled{13} will be used by default if not specified.

\begin{table}[t]
\centering
    \minipage{1\linewidth}
    \vskip -5pt
    \caption{\small Ferret-Bench for referring and grounding VQA, based on Ferret~\citep{you2023ferret} equipped with different image encoders. Models are evaluated with OpenAI \texttt{gpt-4o} API. $^*$replace Ferret visual sampler with \Prompter; Details in Section~\ref{ss_ferret}.
    }
    \label{tab:ferret-bench}
    \centering
    \footnotesize
    \setlength{\tabcolsep}{1pt}
     \renewcommand{\arraystretch}{1}
    \begin{tabular}{lccc|ccc|c}
    \toprule
    Method & ViT & \parbox{1cm}{\centering \scriptsize Region Align} & \parbox{0.95cm}{\centering \scriptsize \#images w/ region labels}  & \parbox{0.85cm}{\centering \scriptsize Ref-Descript.} & \parbox{0.75cm}{\centering \scriptsize Ref-Reason.} & \parbox{0.75cm}{\centering \scriptsize Ground Conv.} & \parbox{0.9cm}{\centering \scriptsize Avg. \\ {\color{magenta} $\Delta$CLIP}}\\
    \midrule
    CLIP & B/16 & None & None & 47.5&	50.3&	45.3&	47.7 \\
    \Ours & B/16 & \scriptsize RoI-Align & 300M & 48.0	&48.4 &40.0	&45.5\\
    \Ours & B/16 & \scriptsize \texttt{Prompter}& 300M & 50.2 &	55.5&	41.5&	49.1\\
    \rowcolor{LightCyan}\Ours & B/16 & \scriptsize \texttt{Prompter} & 2B & 53.6&	53.7& 42.2&	49.8 \scriptsize ({\color{magenta}+2.1})\\
    \rowcolor{LightCyan}\Ours$^*$ & B/16 &  \scriptsize \texttt{Prompter} & 2B & 54.8&	54.9&	44.7&	\textbf{51.5} \scriptsize ({\color{magenta}+3.7}) \\
    \midrule
    \midrule
    \scriptsize OAI-CLIP & L/14 & None & None & 50.8	& 55.4& 45.7&	50.6\\
    CLIP & L/14 & None & None & 54.2&	54.6&	43.3&	50.7\\
    \Ours & L/14 & \scriptsize \texttt{Prompter} & 300M & 51.0&	65.7&	44.9& 53.9 \\
    \rowcolor{LightCyan}\Ours & L/14 & \scriptsize \texttt{Prompter} & 2B & 55.9&	63.3&	46.0&	55.1 \scriptsize ({\color{magenta}+4.4})\\
    \rowcolor{LightCyan}\Ours$^*$ & L/14 & \scriptsize \texttt{Prompter} & 2B & 56.3&	67.4&	47.1&	\textbf{56.9} \scriptsize ({\color{magenta}+6.2})\\
    \bottomrule
    \end{tabular}
    \endminipage
    \vskip -10pt
\end{table}

\begin{table*}[t]
    \centering
    \minipage{1\linewidth}
    \caption{\small Results on referred-LVIS object classification, referring expression comprehension ($0.5$ IoU on RefCOCO, RefCOCO+, RefCOCOg), and phrase grounding ($0.5$ IoU on Flickr30k Entities). \Ferret$^*$: replace visual sampler in \Ferret with \Ours prompter.}
    \label{tab:refcoco}
    \centering
    \footnotesize
    \setlength{\tabcolsep}{0.8pt}
     \renewcommand{\arraystretch}{1.25}
    \begin{tabular}{l|c|ccc|ccc|ccc|cc|cc|c}
    \toprule
    Model & Encoder & \multicolumn{3}{c}{RefLVIS} & \multicolumn{3}{c}{RefCOCO} & \multicolumn{3}{c}{RefCOCO+} & \multicolumn{2}{c}{RefCOCOg} & \multicolumn{2}{c}{Flickr} &  Avg.\\
    & & box & point & \parbox{0.75cm}{\centering free-form} & val & testA & testB & val & testA & testB & val & test & val & test & ({\color{magenta} $\Delta$ to CLIP}) \\
    \midrule
    \Ferret & CLIP B/16 & 72.5&	56.9&	57.2& 80.7&	84.2&	77.1&	71.9&	76.1&	63.7&	75.9&	76.2& 76.2&	78.3 & 72.8\\
    \rowcolor{LightCyan}\Ferret & \Ours B/16 & 74.3&	56.7&	60.2& 84.2&	87.0&	80.0&	74.7&	80.0&	67.0&	78.8&	79.5& 80.0&	81.5& 75.7 ({\color{magenta}+2.9})\\
    \rowcolor{LightCyan}\Ferret$^*$ & \Ours B/16 & 78.9&	58.2&	61.4& 84.4&	86.8&	78.9&	74.0&	78.7&	65.5&	78.0&	78.7& 80.1&	81.4& \textbf{75.8} ({\color{magenta}+3.0})\\
    \midrule
    \midrule
    Shikra~\cite{chen2023shikra} & {\scriptsize OpenAI-CLIP L/14} & 57.8 & 67.7 & n/a & 87.0& 90.6& 80.2& 81.6& 87.4& 72.1& 82.3& 82.2& 75.8& 76.5& -\\
    \Ferret~\cite{zhang2024ferret} & {\scriptsize OpenAI-CLIP L/14} & 79.4&	67.9&	69.8& 87.5&	91.4&	82.5&	80.8&	87.4&	73.1&	83.9&	84.8& 80.4&	82.2& 80.8\\
    \Ferret & CLIP L/14 & 78.7&	66.9&	70.2& 88.0&	90.4&	83.5&	80.1&	85.8&	73.3&	82.8&	83.4& 79.0&	80.1& 80.2\\
    \rowcolor{LightCyan}\Ferret & \Ours L/14 & 81.6&	67.9&	69.9& 89.0&	91.0&	84.7&	81.4&	86.8&	74.7&	84.0&	85.2& 82.3&	83.3& \textbf{81.7} ({\color{magenta}+1.5})\\
    \rowcolor{LightCyan}\Ferret$^*$ & \Ours L/14 & 79.8&	67.9&	69.1& 88.2&	91.1&	84.5&	80.6&	86.7&	73.9&	84.8&	85.1& 82.4&	83.5& 81.4 ({\color{magenta}+1.2})\\
    \bottomrule
    \end{tabular}
    \endminipage
    \vskip -10pt
\end{table*}

\begin{table}[t]
    \minipage{1\linewidth}
    \vskip -5pt
    \caption{\small Results on multimodal benchmarks using LLaVA-1.5 and Open-LLaVA-NeXT with ViT-L/14 and Vicuna-7B.}
    \label{tab:llava}
    \centering
    \small
    \setlength{\tabcolsep}{0.8pt}
     \renewcommand{\arraystretch}{1}
    \begin{tabular}{l|ccccccc}
    \toprule
    Method & {\small LLaVA$^\text{W}$}	& TextVQA & GQA	& MMVet &	POPE	& MMEp & MMEc	\\
    \midrule
    \multicolumn{8}{c}{{\footnotesize \centering LLaVA-1.5}}\\
    \midrule
    CLIP & 59.3 &	53.3&	62.2&	30.0&	86.7&	1451.4&	254.3\\
    \rowcolor{LightCyan}
    \Ours & 64.3 &	54.9&	62.7&	31.5&	87.3&	1482.0&	288.9\\
    \midrule
    \multicolumn{8}{c}{{\footnotesize \centering Open-LLaVA-NeXT}}\\
    \midrule
    CLIP & 67.3	&61.4&	63.5&	38.5&	87.9 &1486.1	&279.6\\
    \rowcolor{LightCyan}
    \Ours & 69.5&	61.9&	64.2&	40.2&	88.3 &1451.1	&312.5\\
    \bottomrule
    \end{tabular}
    \endminipage
    \vskip -10pt
\end{table}

\subsection{Referring and Grounding with Ferret}
\label{ss_ferret}
As discussed in Section~\ref{s_intro}, a key motivation is to provide an enhanced image encoder for training MLLMs, particularly for tasks requiring fine-grained image understanding. A notable example is Ferret~\citep{you2023ferret}, a recently proposed MLLM that builds on LLaVA and aims to handle more advanced spatial interactions, such as referring and grounding in VQA tasks. Ferret can take region prompts such as a box, a point, or a free-form location referring to the input image as input, and answer a question specific to the region such as ``Do you know when the object[region] was invented?'' Ferret thus requires fine-grained image features from the vision encoder for spatial reasoning. 

We evaluate \Ours by replacing the CLIP ViT encoder with our \Ours ViT as a drop-in replacement. We use the official codebase for training the Ferret model. We further consider a variant based on Ferret: the Ferret model implements a spatial-aware visual sampler that samples image features from the region specified in the question. We replace the sophisticated visual sampler with our simple \texttt{Prompter} introduced in~Section \ref{ss_cloc_arch} to extract region embedding with $\vz = \texttt{Prompter}(\vl, f_I'(\vx))$ instead, as illustrated in~\autoref{fig:overview}(right). 

In~\autoref{tab:ferret-bench}, we evaluate different pre-trained image encoders on the Ferret-Bench benchmark~\citep{you2023ferret}. Ferret-Bench includes challenging multimodal dialogue-style VQA of three tasks constructed with GPT-4. Results show that our \Prompter is essential to improve upon the CLIP baseline -- RoI-Align may even slightly degrade performance. Scaling region labels from $300$M to $2$B further improves performance. Interestingly, our \Prompter (denoted as $^*$) can be a replacement of the \Ferret visual sampler in fine-tuning, which is simpler and performs even better up to {\color{magenta}$6\%$} against both the OpenAI-CLIP and our in-house CLIP. We also evaluate \Ours (2B labeled) on other referring and grounding tasks ranging from referring object classification, referring expression comprehension, and phrase grounding across multiple datasets. As summarized in~\autoref{tab:refcoco}, \Ours is also superior evidenced by {\color{magenta}$1\sim 3\%$} improvements in average of $13$ evaluation sets.

\subsection{General VQA with LLaVA-1.5 and LLaVA-NeXT}
\label{ss_llava}
We further show that the \Ours encoder is also competitive against CLIP on general VQA tasks without regression and can even provide performance improvements. We use the Vicuna 7B LLM decoder for two experiments based on LLaVA-1.5 (frozen encoder) and Open-LLaVA-NeXT (unfrozen encoder with AnyRes~\citep{liu2024llavanext} inputs). Since general VQA does not provide spatial referring inputs, we simply replace the ViT in LLaVA. \autoref{tab:llava} summarizes the results. Encouragingly, with our \Ours designs, the improved region-level alignment is also beneficial to some general multimodal benchmarks, as they may also require fine-grained image understanding.

\section{Conclusion}
\label{s_conclusion}
Please see~\autoref{sup-ss-disc} for more discussions where we comment on the limitations, future directions, computation cost, design rationales, etc. 

We tackle a deficiency of CLIP, to make the semantics aligned in the vision space for both image and region level. We propose a new pre-training framework that includes innovations in a new learning formulation that a strong encoder should be easily transformed in the foresee of downstream use of MLLMs. Our encoder creates a new possibility for adapting the features with input prompts of interaction together with MLLMs. To resolve the need for large-scale region-text training data, we carefully design a pseudo-labeling pipeline for visually-enriched and spatially-localized captions. Our pre-trained encoder is essentially a drop-in replacement of CLIP, with competitive image-text performance, and extra capability demonstrated in region-text tasks and VQA tasks with MLLMs.

\section*{Acknowledgment}
We thank Yanghao Li, Felix Bai, and many others for their invaluable help and feedback. 

\bibliography{icml2025}
\bibliographystyle{icml2025}

\clearpage
\newpage
\appendix
\section*{\LARGE Appendix}
\renewcommand{\thesection}{\Alph{section}}
\renewcommand{\thetable}{\Alph{table}}
\renewcommand{\thefigure}{\Alph{figure}}
\renewcommand{\theequation}{\Alph{equation}}
\setcounter{figure}{0}  
\setcounter{table}{0}  

\section{Reproducibility Statement}
We made our best efforts to exhaustively state the implementation details. Training hyper-parameters and model architectures are discussed in Section~\ref{ss_promptable},~\ref{ss_cloc_arch}, and~\ref{ss_setup}, with a summary in~\autoref{sup-s-exp} and~\autoref{tab:hyper}. For evaluation, as mentioned in Section~\ref{ss_setup}, we strictly follow the official setup with the codebase released by the original authors if applicable, with details provided in Section~\ref{sup-ss-eval}. For our datasets, we provide data processing details in Section~\ref{s_data} and example codes in~\autoref{sup-s-vesl}. We are working hard on releasing the annotations with internal approvals.   

\section{Experiment Details}
\label{sup-s-exp}
We provide the omitted experiment details for pre-training and the downstream evaluation tasks.

\subsection{Pre-training Hyper-parameters}
For pre-training both the in-house CLIP baseline and \Ours, we mainly follow the hyper-parameters in~\citep{radford2021learning} to train on our in-house datasets. The training images are identical for CLIP and \Ours, while \Ours is trained on the extra region-text annotations of the same images via the proposed \OWL pipeline (details in Section~\ref{sup-s-vesl}).~\autoref{tab:hyper} summarizes the general training hyper-parameters used for all experiments and the setup for components specific to \Ours.

In terms of the \Ours architecture, as illustrated in~\autoref{fig:arch}, the image and text encoders including the attention pooling and projection layers follow the same as OpenAI-CLIP~\citep{radford2021learning}. Our \Prompter consists of a positional encoding matrix for bounding boxes, and a single-layer single-head transformer encoder with another set of the global average pooler and a projection layer to map the region embeddings into the same dimension as the CLIP text/image embeddings.    

\begin{table*}[h!]
    \minipage{1\linewidth}
    \vskip -10pt
    \caption{\small Pre-training hyper-parameters and settings for the in-house CLIP baseline and \Ours.}
    \label{tab:hyper}
    \centering
    \small
    \setlength{\tabcolsep}{2.5pt}
    \renewcommand{\arraystretch}{1.1}
    \begin{tabular}{l|c}
    \toprule
    \multicolumn{2}{c}{General}\\
    \midrule
    Batch size & $32768$ \\
    Image size & $224\times 224$ (ViT B/16) or  $336\times 336$ (ViT L/14, H/14)\\
    Image pre-processing & long-side resizing with padding (\ie, \texttt{tf.image.resize\_with\_pad})\\
    Text tokenizer & T5~\citep{raffel2020exploring}, lowercase \\
    Text maximum length & $77$ tokens \\
    Steps & $439087$ (\ie, $\sim 14$B examples seen) \\
    Optimizer & AdamW ($\beta_1=0.9, \beta_2=0.98$) \\
    Peak learning rate (LR) & $0.0005$\\
    LR schedule & cosine decays with linear warm-up (first $2$k steps)\\
    Weight decay & $0.2$ \\
    Dropout rate & $0.0$ \\
    \midrule
    \multicolumn{2}{c}{\Ours}\\
    \midrule
    \# of sampled regions & maximum $M=4$ per image \\
    \Ours loss weight & $\lambda=1.0 \times \frac{\# \text{of images contain region text in the mini-batch}}{\text{batch size}}$ (in~\autoref{eq:loss}) \\
    Encoding box prompts & sinusoidal positional encoding of coordinates (top-left and bottom right of a box) \\
    Encoding text prompts & encoded by re-using the text encoder (w/ pooling \& projection)\\
    \Prompter architecture & a single-layer single-head transformer encoder (same feature dimension as the ViT) \\
    \texttt{BoxHead} architecture & 2-layer MLP with GELUs activations \citep{hendrycks2016gelu} \\
    \bottomrule
    \end{tabular}
    \endminipage
    \vskip -10pt
\end{table*}

\subsection{Evaluation Tasks}
\label{sup-ss-eval}
We provide more details about the tasks constructed for evaluating the encoders in Section~\ref{s:exp}. 

\paragraph{Zero-shot region tasks.}
Our \Ours training augments a new capability for CLIP to generate region-level embeddings. This enables us to perform zero-shot region-text tasks, in analogy to the image-text zero-shot tasks like ImageNet classification and COCO text-image retrieval that CLIP has been evaluated on. 

In a similar rationale of image-level evaluation, we further construct region-level tasks including region object recognition and region-text retrieval. For region object recognition, the class names are encoded by the text encoder into class embedding. We do not add the text prompts (\eg, ``a photo of ...'') to object classes used when CLIP~\citep{radford2021learning} evaluated on image classification. The \Ours model takes all the labeled bounding boxes in the images to generate a region embedding $\vz = \texttt{Prompter}(\vl, f_I(\vx))$. The class embedding with the highest similarity is predicted as the class of the region (\ie,  out of $80$ / $1203$ classes for COCO / LVIS). 

For region retrieval, similarly, the \Ours model encodes both the image regions and the region captions from the public region-text GRIT dataset that the regions are annotated by the Kosmos-2 pipeline~\citep{Kosmos2}. We randomly sampled a $2$K image validation set for fast evaluation. We have verified it is statistically stable compared to the whole set that contains about $20$M in total. Unlike image-text retrieval the image captions are likely unique, the objects in regions of many images might be duplicated. Therefore, we opt to report recall@10 rather than recall@1 for GRIT region retrieval in~\autoref{tbl:zero-shot}.    

\paragraph{MLLM tasks.}
To demonstrate our \Ours can benefit MLLM end tasks as a better image backbone, we consider two sets of MLLM experiments. 

First, we experiment with the~\Ferret MLLM that is capable of taking spatial referring inputs for grounding and referring VQA tasks\footnote{We use the official Ferret codebase: \url{https://github.com/apple/ml-ferret}.}. \Ferret can consume a point, a bounding box, or a free-form referring. It designs a quite complicated visual sampler module that involves point sampling and $k$NN grouping. We suggest the readers refer to Figure 3 and Section 3.2 in~\citep{you2023ferret} for more details. Here we consider two variants of use cases of \Ours compatible with \Ferret: (1) we only take the ViT encoder in \Ours to replace the CLIP ViT and still use the original \Ferret visual sampler or (2) we further replace the visual sampler with our simple \Prompter (essentially a lightweight transformer encoder with box positional encodings) in Section~\ref{ss_cloc_arch} as illustrated in~\autoref{fig:overview}(3b). More specifically, we simply convert all types of spatial referring as boxes. As evidenced by~\autoref{tab:ferret-bench} and~\autoref{tab:refcoco}, our \Prompter can indeed be a much simpler alternative and may perform even better as it is more consistent with \Ours pre-training. 

Second, we evaluate on general VQA tasks that do not consider extra spatial referring inputs. The pre-trained ViT of \Ours is a drop-in-replacement of CLIP ViT in two sets of experiments of LLaVA-1.5~\citep{liu2023llava} and LLaVA-NeXT~\citep{liu2024llavanext}. The main difference includes different supervised fine-tuning (SFT) sets. Also, LLaVA-NeXT uses the AnyRes technique that decomposes an image into several subimages that are encoded independently with the ViT and concatenated together as the input for the decoder. LLaVA-1.5 by default freezes the ViT while LLaVA-NeXT fine-tunes all parameters during SFT. Since the official LLaVA-NeXT is trained on some proprietary datasets that are not reproducible, we use the Open-LLaVA-NeXT repository\footnote{https://github.com/xiaoachen98/Open-LLaVA-NeXT}. Our experiments in~\autoref{tab:llava} demonstrate \Ours not only slightly improves such general VQA besides \Ferret tasks but also generalizes well for both LLaVA-1.5 and LLaVA-NeXT settings.

\section{\OWL Data Engine}
\label{sup-s-vesl}

We provide more information about our pseudo-labeling data pipeline proposed in Section~\ref{s_data}. 

\subsection{Implementation Details}
\label{sup-ss-vesl-ner}
As already mentioned in Section~\ref{s_data}, there are three steps for \OWL: image re-captioning, region phrase candidates extraction from the captions, and open-vocabulary (OV) detection given the region candidates as queries. 

For the re-captioning, the goal is to replace AltText with long, diverse, and detailed captions that can be used to generate more visual concepts as the region candidate phrases for the OV detector. Technically, any strong image captioner can be an option. In our paper, we adopt the VeCap pipeline~\citep{lai2023scarcity} and leverage their images with enriched captions.

To extract region phrase candidates from the long captions, we adopt name entity recognition (NER) to extract leaf entities from the captions, inspired by~\citep{zhang2022glipv2}. The code listing below shows the Python example implementation, where stop-words and common generic words are filtered, following~\citep{minderer2024scaling}. 

Generating bounding boxes and assigning region captions can be done by querying an OV objection detector. We adopted the OWLv2 detector~\citep{minderer2024scaling} with their pre-trained L/14 checkpoint\footnote{OWLv2 CLIP L/14 ST+FT in: https://github.com/google-research/scenic/tree/main/scenic/projects/owl\_vit} to annotate inputs with $448\times 448$ image resolutions.

\begin{lstlisting}[language=Python, caption=Python example codes for Step 2 of VESL in Section 4 for extracting text candidate queries from a caption.]
from typing import Iterable, List
import nltk

# STOPWORDS_EN and COMMON_GENERIC_WORDS are following:
# Section A.2 (Minderer et al., 2024) 

# Stopwords from nltk.corpus.stopwords.words("english"):
STOPWORDS_EN = frozenset({
    "a", "about", "above", "after", "again", "against", "all", "am", "an",
    "and", "any", "are", "as", "at", "be", "because", "been", "before", "being",
    "below", "between", "both", "but", "by", "can", "did", "do", "does",
    "doing", "don", "down", "during", "each", "few", "for", "from", "further",
    "had", "has", "have", "having", "he", "her", "here", "hers", "herself",
    "him", "himself", "his", "how", "i", "if", "in", "into", "is", "it", "its",
    "itself", "just", "me", "more", "most", "my", "myself", "no", "nor", "not",
    "now", "of", "off", "on", "once", "only", "or", "other", "our", "ours",
    "ourselves", "out", "over", "own", "s", "same", "she", "should", "so",
    "some", "such", "t", "than", "that", "the", "their", "theirs", "them",
    "themselves", "then", "there", "these", "they", "this", "those", "through",
    "to", "too", "under", "until", "up", "very", "was", "we", "were", "what",
    "when", "where", "which", "while", "who", "whom", "why", "will", "with",
    "you", "your", "yours", "yourself", "yourselves"
})

# These words were found by manually going through the most common 1000 words
# in a sample of alt-texts and selecting generic words without specific meaning:
COMMON_GENERIC_WORDS = frozenset({
    "alibaba", "aliexpress", "amazon", "available", "background", "blog", "buy",
    "co", "com", "description", "diy", "download", "facebook", "free", "gif",
    "hd", "ideas", "illustration", "illustrations", "image", "images", "img",
    "instagram", "jpg", "online", "org", "original", "page", "pdf", "photo",
    "photography", "photos", "picclick", "picture", "pictures", "png", "porn",
    "premium", "resolution", "royalty", "sale", "sex", "shutterstock", "stock",
    "svg", "thumbnail", "tumblr", "tumgir", "twitter", "uk", "uploaded", "vector",
    "vectors", "video", "videos", "wallpaper", "wallpapers", "wholesale", "www",
    "xxx", "youtube"
})


def _is_all_stopwords(query_words: Iterable[str]) -> bool:
    return set(query_words).issubset(STOPWORDS_EN)


def _get_name_entities(words: List[str]) -> List[str]:
    """
    Returns name entities of image caption as queries, similar to GLIP.
    """
    pos_tags = nltk.pos_tag(words)
    grammar = "NP: {<DT>?<JJ.*>*<NN.*>+}"
    cp = nltk.RegexpParser(grammar)
    result = cp.parse(pos_tags)

    queries = []
    for subtree in result.subtrees():
        if subtree.label() == "NP":
            query_words = [t[0] for t in subtree.leaves()]
            # Don't use it if it only consists of stop words.
            if _is_all_stopwords(query_words):
                continue
            queries.append(" ".join(query_words))
    return queries


def find_noun_phrases(
    caption: str, max_num_queries: int = 20,
) -> List[str]:
    caption = caption.lower()
    tokens = nltk.word_tokenize(caption)
    # Remove common generic words.
    words = [w for w in tokens if w not in COMMON_GENERIC_WORDS]
    queries = _get_name_entities(words)[:max_num_queries]
    return queries

candidate_quries = find_noun_phrases(caption) 
\end{lstlisting}

\subsection{More Visualizations}
\begin{figure*}[h]
    \centering
    \vskip-15pt
    \includegraphics[width=0.6\linewidth]{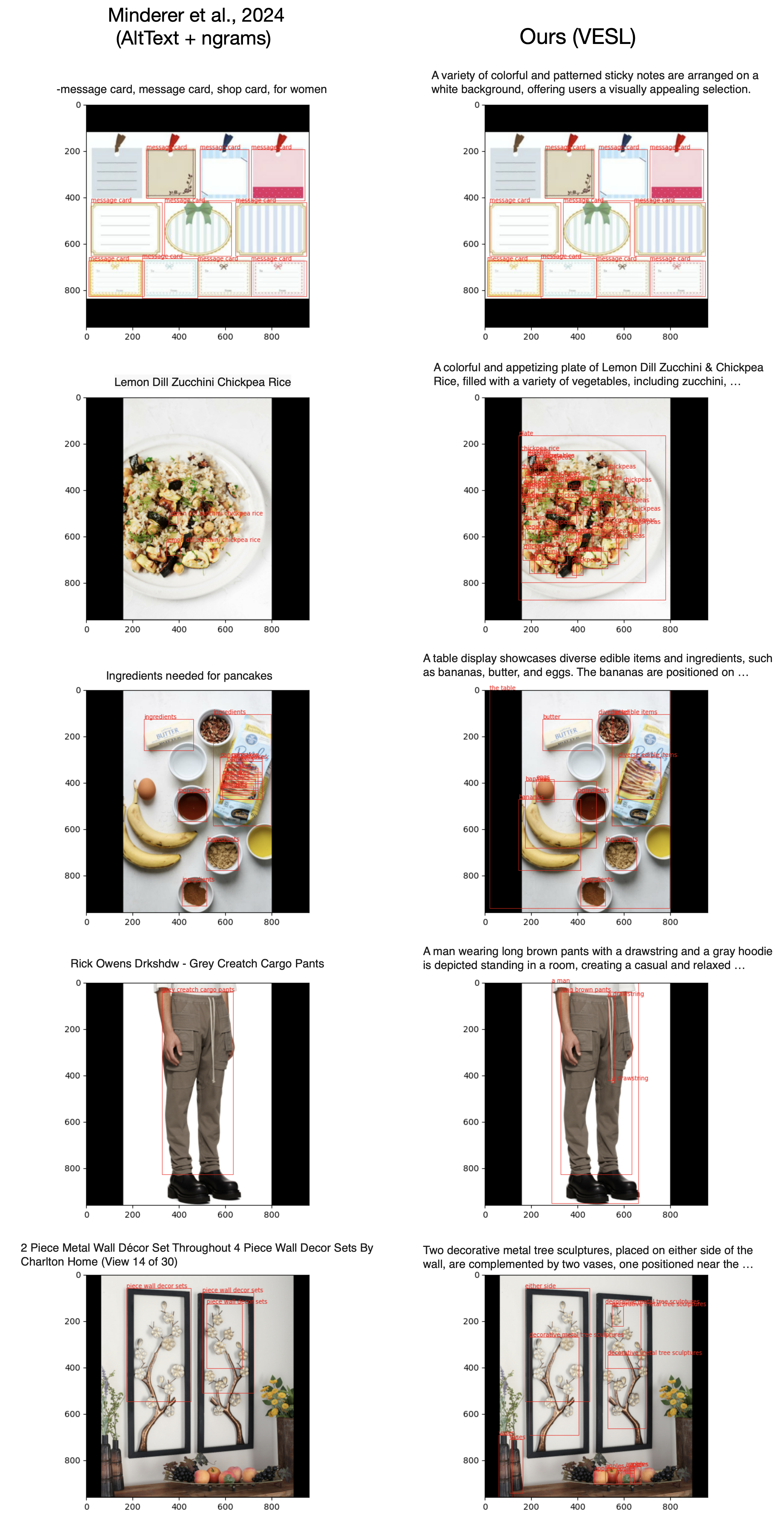}
    \caption{\small Examples comparing our \OWL and the labeling approach in~\citep{minderer2024scaling} that directly uses the $n$-grams of the crawled AltText. For \OWL, each image is annotated with the visual-enriched caption to replace the AltText, which is used to generate region text candidates that capture the image content better.}
    \label{fig:vis}
    \vskip -5pt
\end{figure*}

As mentioned in the remarks of Section~\ref{s_data}, we found the AltText sourced from the original web-crawled images might not have enough details describing the subimage content, thus limiting the diversity and quality of the text candidate queries for the OV detector to detect more meaningful objects. In~\autoref{fig:vis} we show some cherry-picked examples (since the web-crawled images are quite noisy) just to demonstrate the reasons why high-quality captions can help our region-text annotation pipeline. In~\citep{minderer2024scaling}, the queries are generated by the $n$-grams of the AltText, while ours are by NER as described in Section~\ref{sup-ss-vesl-ner} on top of the visually-enriched re-captions. Note that, in both methods we use the same pre-trained OV detector but with different approaches to generate the queries.

As shown in~\autoref{fig:vis}, for easier images like the first row, both methods are doing reasonably well to detect ``message card''. However, when the scene becomes complicated (\eg, the second row), our methods can detect more objects since more visual concepts can be extracted from our rich caption as queries for the detector. 
Similarly, it can be seen that our method captures more items that the AltText missed, \eg, ``banana'', ``eggs'', ``butter'', etc in the third row; ``drawstring'' in the fourth row; ``apples'' and ``vases'' in the last row. Also, it is more likely to extract a more detailed description of the region rather than a class name, such as ``green-roofed cottage nestles'' in~\autoref{fig:vesl} and ``decorative metal tree sculptures'' in the image in the last row of ~\autoref{fig:vis}. We believe such high-quality region labels essentially contribute to better supervision for \Ours pre-training.

\section{More Discussions}
\label{sup-ss-disc}

\paragraph{Limitations.}
One limitation for \Ours is the labeling efforts in preparing the training data. As we discussed in Section~\ref{s_intro}, there are no public large-scale region-text datasets since it is expensive to infer such labels up to the scales we consider here. Unlike previous work~\citep{zhong2022regionclip} that cropping boxes from images for annotating, our \OWL inference in \emph{image-level} thus the cost does not scale with the number of detected regions. With that being said, such inference still requires hundreds of GPUs running in parallel for days to scale up to billions of images. We are working on releasing the annotations to accelerate future research for the community. 

For \Ours, we focus on the training objective and framework formulation, while making minimal efforts on hyper-parameter tuning, architecture search, dataset cleaning, and etc., thus better performance could be achieved. Besides, although we have included extensive standard evaluation tasks, the fine-grained region knowledge could also be useful on more other under-explored tasks.

\paragraph{Future directions.} We suggest promising future directions. In Section~\ref{ss_promptable}, our \Prompter formulation can take flexible prompts to guide the embeddings for specific tasks. In this work, we consider a prompt as a single bounding box or a text caption, but it has the potential to expand to various types such as points, a mask, users' free-form referring, or multiple prompts in multiple types together. We think a more versatile \Prompter with co-designs for different objectives can have a big potential. Similarly, our \OWL labeling pipeline limits to detection box format. Annotators supported for more formats may further boost it. We believe our approach is promising, as more attention has been drawn recently for better re-captions~\citep{li2024if,laclip} that \OWL relies on. In addition, \Ours model provides a new capability to extract region features without further training, and thus can be used as a foundation model for exploring new VL applications.

\paragraph{Training cost.} We comment on the computation cost of our framework. 
Our large models (ViT L/14) were trained on $1024$ v5p TPUs for about $6$ days. To optimize~\autoref{eq:cloc}, \Ours needs extra computation. The main overheads come from the contrastive matrix but not the lightweight \Prompter. Fortunately, we found it feasible since (1) only a few boxes in each image need to be sampled per update; (2) the loss computation becomes a smaller proportion when the ViT scales up. Overall, we found the computation acceptable compared to CLIP. More memory-efficient optimization like SigLIP~\citep{zhai2023sigmoid} can be implemented with JAX \texttt{shard\_map}\footnote{https://jax.readthedocs.io/en/latest/jep/14273-shard-map.html} ops. 

\paragraph{Discussions on design rationals.}
Besides the main discussions we have stressed in the main text, here we provide more thoughts behind our design rationales that a reader may be wondering.

\emph{(1) Why not use a local-enhanced encoder?} We would like to note that many encoders with great localization like DINOv2~\citep{oquab2023dinov2}, OWLv2~\citep{minderer2024scaling}, CLIP-Self~\citep{wu2023clipself}, etc. are developed specifically for dense vision tasks that cannot perform image zero-shot tasks like CLIP and \Ours. We would like to emphasize that our goal is to build a drop-in-replacement of CLIP encoder with better localization, without sacrificing CLIP's original capabilities such as image zero-shot tasks and its important backbone position for MLLMs. Furthermore, perhaps well-known within the MLLMs community, these encoders have been shown in recent reports that they are not comparable enough to compete with CLIP as the vision backbone for MLLM tasks~\citep{tong2024cambrian} due to CLIP's superiority in vision-language alignment. We thus believe enhancing CLIP itself is more demanding as this paper focuses on. 

\emph{(2)  Why not just train a CLIP with object detection?} One may wonder why we do not just train an encoder with joint optimization of the CLIP contrastive loss with some object detection loss instead of the \Ours design of~\autoref{eq:loss}. 

Although it sounds like a plausible approach, we would like to point out that contrastive pre-training and object detection are fundamentally quite different in their technical rationales. CLIP pre-training is often on large batches of low-resolution and noisy images, while object detection is trained on small batches of high-resolution images. CLIP is by default trained from scratch and object detection is typically initialized from pre-trained encoders and focuses on the detection head. Furthermore, detection requires heavy computation on box proposals to detect all boxes appearing in an image, while our region-text contrastive design allows us to flexibly sample fewer regions per image as motivated in~\autoref{eq:grounding_loss}. Overall, their data pipeline and distributed training setup are not on the same scale thus such joint training may not be very reasonable. 

With that being said, some previous works do have attempts that are the exceptions but only for some but not all of the mentioned aspects, and mainly for the purpose of detection. For instance, DetCLIP-v2~\citep{yao2023detclipv2} adds image-text contrastive loss into detection loss to improve open-vocabulary capability for detection. OWLv2 pre-trains the detector with rather small resolutions but still with a batch size of a maximum $256$ since each image will need to predict up to $100$ boxes during training. Both DetCLIP-v2 and OWLv2 fine-tune from a pre-trained encoder. 

On the contrary, we study pre-training the encoder from scratch, which may be complementary to the previous efforts. \Ours maximizes the similarity in co-design with CLIP, thus making it much easier to develop within the same codebase. 

\emph{(3) Do we really need to train \Ours from scratch? What if we fine-tune from CLIP?} As CLIP pre-training is expensive, one may wonder if it is necessary to train from scratch on the proposed region-text datasets, or if we can initialize from a standard CLIP trained on image-text pairs only and fine-tunes with \Ours for a shorter stage. Our early investigation, even with extensive hyper-parameter tuning, suggests it is likely to be suboptimal compared to training from scratch directly. For instance, we initialize from the CLIP model \circled{2} in~\autoref{tbl:zero-shot} and fine-tunes it for another extra $100$K steps with the \Ours training loss~\autoref{eq:loss}. The model reaches $64.1\%/19.1\%$ mAcc on COCO/LVIS region recognition, which is much worse than $70.1\%/27.2\%$ of the trained-from-scratch model \circled{8}, even with more overall training steps.


\end{document}